\documentclass[conference]{IEEEtran}
\ifCLASSINFOpdf
\else
\fi
\usepackage{amsfonts}
\usepackage{amsmath}
\usepackage{enumitem} 
\usepackage{subfig}
\usepackage{color,soul}
\usepackage{graphicx}
\usepackage{cite}
\usepackage{balance}

\begin{document}
%
% paper title
% Titles are generally capitalized except for words such as a, an, and, as,
% at, but, by, for, in, nor, of, on, or, the, to and up, which are usually
% not capitalized unless they are the first or last word of the title.
% Linebreaks \\ can be used within to get better formatting as desired.
% Do not put math or special symbols in the title.
\title{Real-time Lane Marker Detection Using Template Matching with RGB-D Camera}

% author names and affiliations
% use a multiple column layout for up to three different
% affiliations
\author{\IEEEauthorblockN{Cong Hoang Quach, Van Lien Tran, Duy Hung Nguyen, Viet Thang Nguyen, \\Minh Trien Pham and Manh Duong Phung}
\IEEEauthorblockA{VNU University of Engineering and Technology\\
Hanoi, Vietnam\\
Email: hoangqc@vnu.edu.vn}
}

% conference papers do not typically use \thanks and this command
% is locked out in conference mode. If really needed, such as for
% the acknowledgment of grants, issue a \IEEEoverridecommandlockouts
% after \documentclass

% for over three affiliations, or if they all won't fit within the width
% of the page, use this alternative format:
% 
%\author{\IEEEauthorblockN{Michael Shell\IEEEauthorrefmark{1},
%Homer Simpson\IEEEauthorrefmark{2},
%James Kirk\IEEEauthorrefmark{3}, 
%Montgomery Scott\IEEEauthorrefmark{3} and
%Eldon Tyrell\IEEEauthorrefmark{4}}
%\IEEEauthorblockA{\IEEEauthorrefmark{1}School of Electrical and Computer Engineering\\
%Georgia Institute of Technology,
%Atlanta, Georgia 30332--0250\\ Email: see http://www.michaelshell.org/contact.html}
%\IEEEauthorblockA{\IEEEauthorrefmark{2}Twentieth Century Fox, Springfield, USA\\
%Email: homer@thesimpsons.com}
%\IEEEauthorblockA{\IEEEauthorrefmark{3}Starfleet Academy, San Francisco, California 96678-2391\\
%Telephone: (800) 555--1212, Fax: (888) 555--1212}
%\IEEEauthorblockA{\IEEEauthorrefmark{4}Tyrell Inc., 123 Replicant Street, Los Angeles, California 90210--4321}}

% use for special paper notices
%\IEEEspecialpapernotice{(Invited Paper)}

% make the title area
\maketitle

% As a general rule, do not put math, special symbols or citations
% in the abstract
\begin{abstract}
This paper addresses the problem of lane detection which is fundamental for self-driving vehicles. Our approach exploits both colour and depth information recorded by a single RGB-D camera to better deal with negative factors such as lighting conditions and lane-like objects. In the approach, colour and depth images are first converted to a half-binary format and a 2D matrix of 3D points. They are then used as the inputs of template matching and geometric feature extraction processes to form a response map so that its values represent the probability of pixels being lane markers. To further improve the results, the template and lane surfaces are finally refined by principal component analysis and lane model fitting techniques. A number of experiments have been conducted on both synthetic and real datasets. The result shows that the proposed approach can effectively eliminate unwanted noise to accurately detect lane markers in various scenarios. Moreover, the processing speed of 20 frames per second under hardware configuration of a popular laptop computer allows the proposed algorithm to be implemented for real-time autonomous driving applications.
\end{abstract}

% no keywords

% For peer review papers, you can put extra information on the cover
% page as needed:
% \ifCLASSOPTIONpeerreview
% \begin{center} \bfseries EDICS Category: 3-BBND \end{center}
% \fi
%
% For peerreview papers, this IEEEtran command inserts a page break and
% creates the second title. It will be ignored for other modes.
\IEEEpeerreviewmaketitle

\section{Introduction}
Studies on automated driving vehicles have received much research attention recently due to rapid advancements in sensing and processing technologies. The key for successful development of those systems is their perception capability, which basically includes two elements: road and lane perception and obstacle detection. It is certainly that road boundaries and lane markers are designed to be highly distinguishable. Those features however are deteriorated over time under influences of human activities and weather conditions. They together with the occurrence of various unpredictable objects on roads cause the lane detection a challenging problem. Studies in the literature deal with this problem by using either machine learning techniques or bottom-up features extraction. 

In the first approach, data of lanes and roads is gathered by driving with additional sensors such as camera, lidar, GPS and inertial measurement unit (IMU) \cite{Janai2017}. Depending on the technique used, the data can be directly fed to an unsupervised learning process or preprocessed to find the ground truth information before being used as inputs of a supervised learning process. In both cases, advantages of scene knowledge significantly improve the performance of lane and road detection. This approach however faces two main drawbacks. First, it requires large datasets of annotated training examples which are hard and costly to build. Second, it lacks efficient structures to represent the collected 3D data for training and online computation. As those data are usually gathered under large-scale scenes and from multiple cameras, current 3D data structures such as TSDF volumes \cite{Curless1996}, 3D point clouds \cite{Phung2016}, or OctNets \cite{Riegler2017} are highly memory-consuming for real-time processing.

In the bottom-up feature extraction approach, low-level features based on specific shapes and colours are employed to detect lane markers \cite{Hillel2012}. In \cite{Samadzadegan2006,Nieto2008}, gradients and histograms are used to extract edge and peak features. In \cite{McCall2006}, steerable filters are introduced to measure directional responses of images  using convolution with three kernels. The template matching based on a bird’s-eye view transformed image are proposed in \cite{BorkarACIVS2010} to improve the robustness of lane detection. Compared with machine learning, the feature-based approach requires less computation and smaller datasets. The detection results however are greatly influenced by lighting conditions, intensity spikes and occluded objects \cite{Samadzadegan2006,Nieto2008, BorkarACIVS2010}.

On another note, both aforementioned approaches mainly rely on colour (RGB) images. The depth (D) information however has not been exploited. In lane marker detection, using both depth and colour information can dramatically increase the accuracy and robustness of the estimation tasks, e.g, obstacles like cars and pedestrians can be quickly detected and rejected by using depth information with a known ground model. The problem here is the misalignment between the colour and depth pixels which are recorded by different sensors such as RGB camera and lidar with heterogeneous resolutions and ranges. With recent advance in sensory technology, this problem can be handled by using RGB-D sensors such as Microsoft Kinect and Intel Realsense SR300 \cite{Han2013, Fankhauser2015, Dinh2014}. Those sensors can provide synchronised RGB-D streams at a high frame rate in both indoor and outdoor environments by using structured-light and time-of-flight technologies.

In this paper, we present a novel method for lane boundaries tracking using a single RGB-D camera. Low-level features of land markers are first extracted by using template matching with enhancements from geometric features. Dynamic thresholds are then applied to obtains the lane boundaries. Here, our contributions are threefold: (i) the formulation of a respond map for lane marker by using both colour and depth information; (ii) the proposal of a processing pipeline and refining feedback for RGB-D template matching; and (iii) the creation of 3D lane model estimation method by using high reliable lane marker points to deal with overwhelmed outlying data in scenes. 

The remaining parts of the paper are structured as follows. Section \ref{method} describes the methodology. Section \ref{experiment} presents the experimental setup and results. The paper ends with conclusions and discussions presented in section \ref{conclusion}.
 
\section{Methodology} \label{method}
\begin{figure}
	\includegraphics[width=8.5cm]{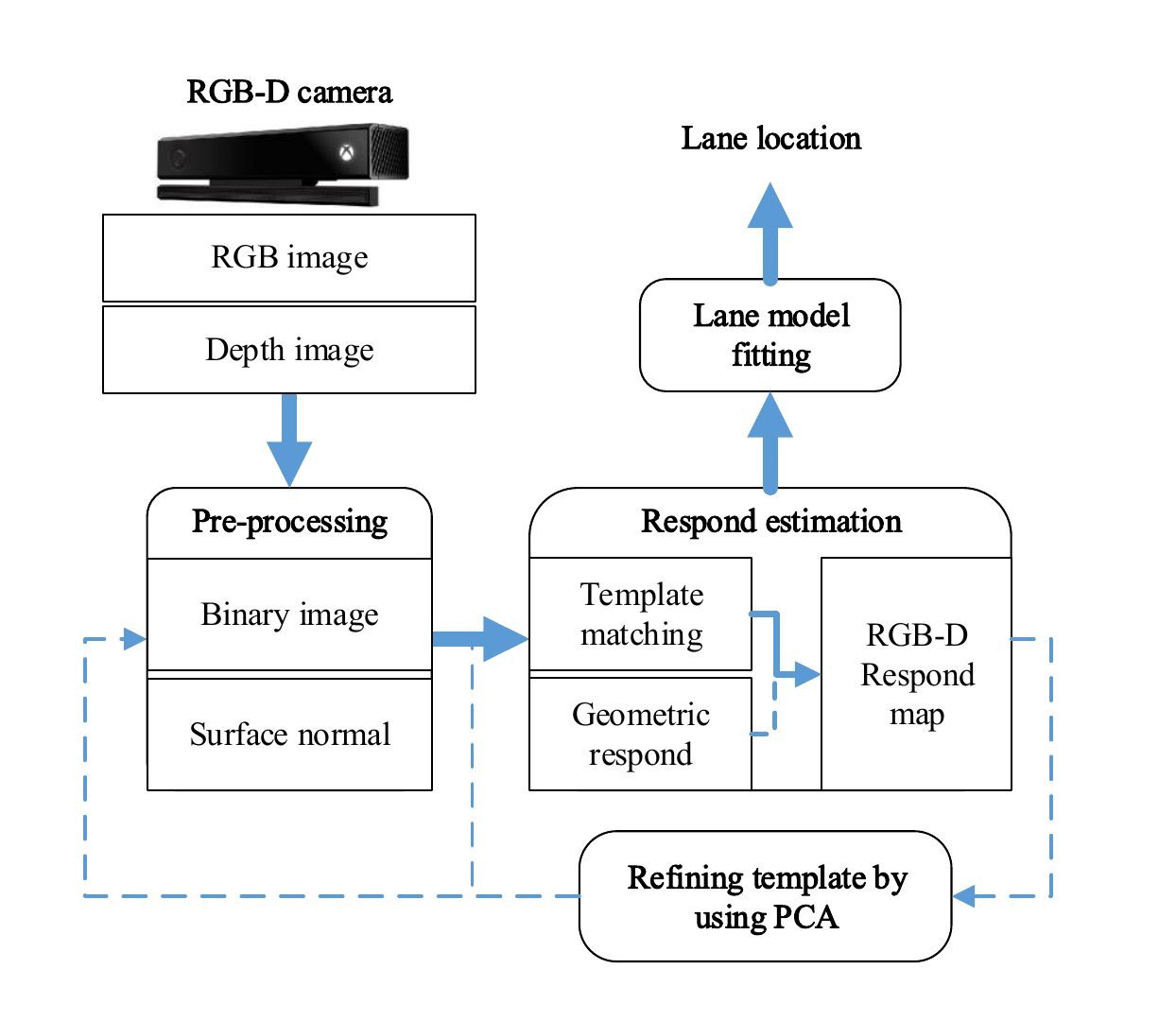}
	\caption{Lane detection pipeline.}
	\label{fig:diagram1}
\end{figure}

An overview of the proposed lane detector is shown in Fig.\ref{fig:diagram1}. A single RGB-D camera attached to the vehicle is used to collect data of the environment. Recorded RGB images are then converted to binary images whereas depth ones are registered and transformed into 3D point clouds. Respond maps of lane markers are then built based on the combination of template matching and geometric feature outputs. The principal component analysis (PCA) technique is then used to refine the templates used. Finally, lane locations are obtained based on its model with a set of detected feature points. Details of each stage are described as follows.

\subsection{Image pre-processing}
In this stage, data from RGB channels are combined and converted to a eight-bit, grayscale image. This image is then converted to a half-binary format by using a threshold $\tau_c$  so that the intensity of a pixel is set to zero if its value is smaller than $\tau_c$. At the same time, data from depth channel is transformed to a 2D matrix of 3D points in which the coordinate of each point, $p=(x,y,z)^T$, is determined by:
\begin{equation}
\label{Eq1}
\left\{
\begin{array}{ll}
x = \frac{i-c_x}{f_x}D(i,j)
\\
y = \frac{j-c_y}{f_y}D(i,j)
\\
z = D(i,j)
\end{array} \right.
\end{equation}
where $D(i,j)$ is the depth value at location $(i,j)$ of the 2D matrix and $(c_x,c_y)$ and $(f_x,f_y)$ are the center and focal length of the camera, respectively. The Fast Approximate Least Squares (FALS) method is then employed to obtain 3D surface normals \cite{Badino2011}. It includes three steps: identifying neighbours, estimating normal vectors based on those neighbours, and refining the direction of the obtained normal vectors. Specifically, a small rectangular window of size $k_N = w\times h$ around the point to be estimated is first determined. Least squares are then formulated to find the plane parameters that optimally fit the surface of that window. The optimisation process uses a loss function defined based on spherical coordinates to find the plane parameters in the local area as:
\begin{equation}
\hat{e} = \sum_{i=1}^{k} (v_i^T\hat{n}-r_i^{-1})^2,
\end{equation}
where $v_i$ is the unit vector, $r_i$ is the range of point $p_i$ in the window $k_N$, and $\hat{n}$ is the normal vector.  $\hat{n}$  is computed by:
\begin{equation}
	  \hat{n}=\hat{M}^{-1}\hat{b},
\end{equation}
where $\hat{M}=\sum_{i=1}^{k} {v_i}{v_i^T}$ and $\hat{b}=\sum_{i=1}^{k}\frac{v_i}{r_i}$ . As matrix  $\hat{M}^{-1}$  only depends on parameters of the camera, it can be pre-computed to reduce the number of multiplications and additions required for computing surface normals.

\subsection{Respond map computation}
Given the standardised colour and depth images, our next step is to compute for each pixel a probability that it belongs to the lane marker. A combination of all probabilities forms a map called \textit{respond map}. For this task, the evaluation is first carried out separately for the colour and depth images. A rule is then defined to combine them into a single map. 

\begin{figure}
	\subfloat[]{\includegraphics[width=1.7in]{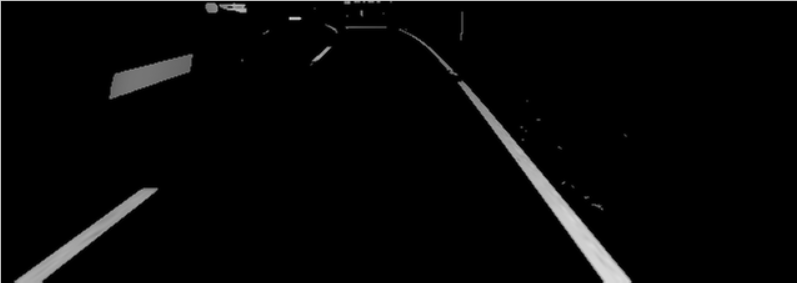}%
		\label{fig:diagram2a}}\vspace{0.1in}
	\subfloat[]{\includegraphics[width=1.7in]{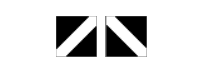}%
		\label{fig:diagram2b}}
	\hspace{0mm}
	\subfloat[]{\includegraphics[width=1.7in]{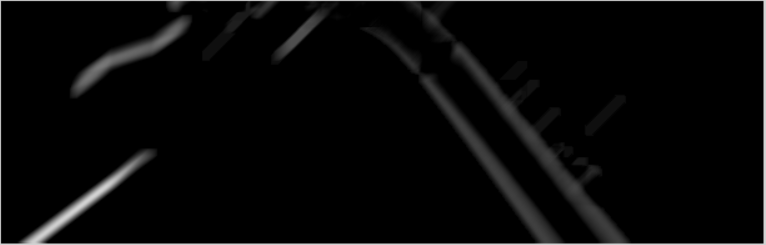}%
		\label{fig:diagram2c}}\vspace{0.1in}
	\subfloat[]{\includegraphics[width=1.7in]{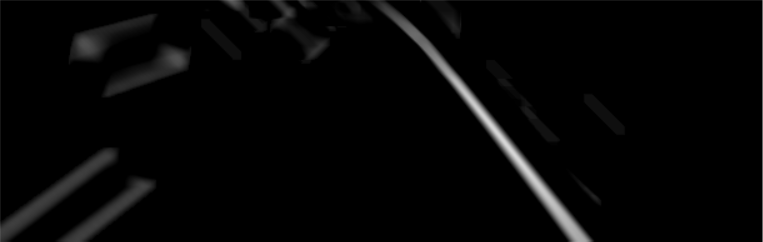}%
		\label{fig:diagram2d}}
	\caption{Template matching process: (a) Haft-binary image; (b) Left and right templates; (c) Matching result with left template; (d) Matching result with right template.}
	\label{fig:diagram2}
\end{figure}

For colour images, we define two templates having shapes similar to the size and direction of lane markers in existing roads as shown in Fig.\ref{fig:diagram2b}. Those templates are then used to extract features of lane markers from half-binary images by using normalized cross correlation (NCC). The matching result, M, as shown in Fig.\ref{fig:diagram2c} and \ref{fig:diagram2d} is normalized to the range from 0 to 1 in which the higher value implies a higher probability of being lane markers.

On the other hand, the geometric feature map, G, is created from the depth image based on a predefined threshold, $T_D$, as:
\begin{equation}
\label{Eq2}
G(i,j) = \left\{ \begin{array}{ll}
\alpha(\vec{n}.\vec{O_y}) +{\beta}\frac{D(i,j)}{T_D} &\mbox{ if $D(i,j)\le T_D$}
\\
\alpha(\vec{n}.\vec{O_y})+{\beta}\frac{j}{imgHeight}&\mbox{ otherwise}
\end{array} \right.
\end{equation}
Eq.\ref{Eq2} can be illustrated as follows:
\begin{itemize}
\item	If the depth value of a pixel is smaller than $T_D$, the corresponding value in G is the dot product between pixel normal $\vec{n}$ and the unit vector $\vec{O_y}$ of the camera view, which has a similar direction as the road's plane normal.
\item	If the depth value is greater than $T_D$ or unknown due to noise, the corresponding value in G is set to a value between 0 and 1 depending on its horizontal location $j$ in the 2D image.
\end{itemize}
Based on the matching result M and the geometric feature map G, the respond map is established by the following equation:
\begin{equation}
\label{Eq3}
R(i,j) = \left\{ \begin{array}{rl} M(i,j) &\mbox{ if $M(i,j)<\tau_G$} \\
M(i,j)+G(i,j) &\mbox{ otherwise}
\end{array} \right.
\end{equation}
where $\tau_G$ is the threshold determined so that $G$ only supports high-reliable colour features in the matching result $M$. Through this response map, both colour and depth information are exploited to evaluate the probability of a pixel belonging to lane markers. 

\begin{figure}
	\subfloat[]{\includegraphics[width=1.7in]{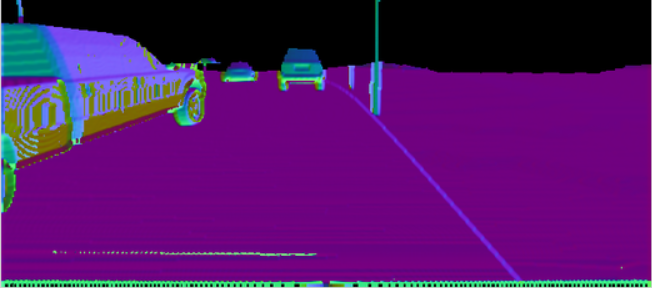}%
		\label{fig:diagram3a}}\vspace{0.1in}
	\subfloat[]{\includegraphics[width=1.7in]{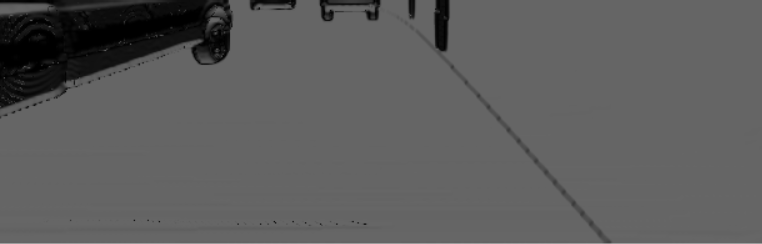}%
		\label{fig:diagram3b}}
	\hspace{0mm}
	\subfloat[]{\includegraphics[width=1.7in]{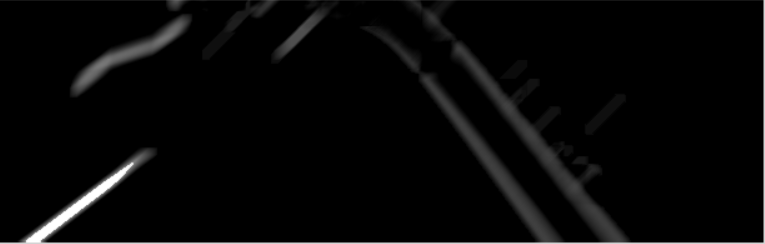}%
		\label{fig:diagram3c}}\vspace{0.1in}
	\subfloat[]{\includegraphics[width=1.7in]{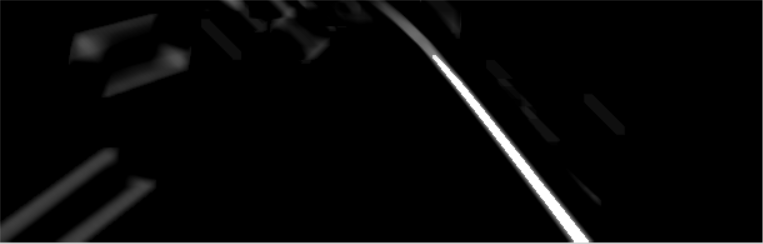}%
		\label{fig:diagram3d}}
	\caption{Computation of respond maps: (a) 3D normal image; (b) G map; (c) respond map of left marker; (d) respond map of right marker.}
	\label{fig:diagram3}
\end{figure}

\begin{figure}
	\centering
	\subfloat[]{\includegraphics[width=3in]{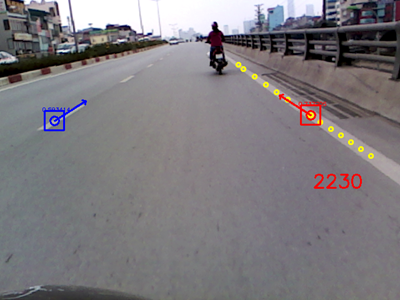}%
		\label{fig:diagram4a}}\vspace{0.1in}
	\subfloat[]{\includegraphics[width=3in]{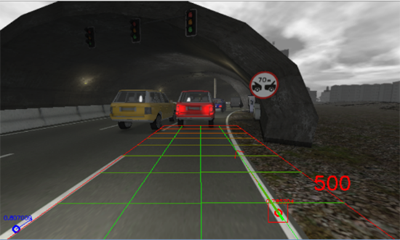}%
		\label{fig:diagram4b}}
	\caption{Template enhancement: (a) Detection result with real datasets in which red and blue rectangles indicate two PCA-based analysis regions, arrows indicate lane direction, and yellow circles represent sliding box centers; (b) Detection result with synthetic datasets in which the grid represents the 3D plane model estimated by our algorithm.}
	\label{fig:diagram4}
\end{figure}

\subsection{Template enhancement}
In lane marker detection, one important issue that need be tackled is the variance of markers with the scale and rotation of the camera. We deal with this problem by applying the principal component analysis (PCA) on the region of R corresponding to the highest probability of being lane markers. This region is selected by first choosing the pixel with the highest value (probability) and then expanding to its surrounding based on threshold ${P_{PCA}}$. The result is a set of points $P_r\subset R$ used as inputs for the PCA. As a result of PCA, the primary eigenvector output forms a new template angle $\theta$ that can be used to adjust the deflecting angle of the template in next frames for better detection.
 
On the other hand, the connected components of lane markers are selected by using a sliding box in the respond map $M$. The box has the same size as the matching templates and the centroid to be the highest positive point. To slide the box, its new origin ${O_b}^i$ is continuously updated from the centroid of the previous subset points $P_r$ as: 
\begin{equation}
\label{Eq4}
O_b^{i+1} = \left\{ \begin{array}{rl}
{O_b}^i + r(\cos \theta, \sin \theta) &\mbox{ if $({O_b}^i-{O_b}^{i+1})^2\le r^2$} \\
\textnormal{centroid of } P_{PCA} &\mbox{ otherwise}
\end{array} \right.
\end{equation}
The stopping criteria include two cases: (i) the origin is out of the image area; or (ii) the set $P_{PCA}$ is null. The main advantage of this method is that it does not require any change in viewpoint procedures as in \cite{BorkarACIVS2010} so that it is less sensitive to noise.

\subsection{Lane model fitting}
The plane model of lanes is defined by three points: two highest-value points in the respond map, $v_{a1}$ and $v_b$  , and the furthest centroid point of the right lane marker, $v_{a2}$ . The plane normal of a lane is defined by the following equation:
$$\hat{n}=\overrightarrow{(v_b - v_{a1})} \times \overrightarrow{(v_{a2}-v_{a1})}$$
It is worth noting that the least square methods like RANSAC are not necessary to use here as most outliners have been removed by our RGB-D matching in previous steps as shown in Fig.\ref{fig:diagram3}. As a result, our method has low computation cost and can overcome the problem relating to quantization errors of the depth map as described in \cite{Han2013}.

\section{Experiment} \label{experiment}
\begin{figure}
	\centering
	\subfloat[]{\includegraphics[width=1.7in]{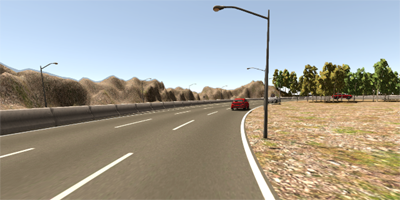}%
		\label{fig:diagram8a}}
	\vspace{0.1in}
	\subfloat[]{\includegraphics[width=1.7in]{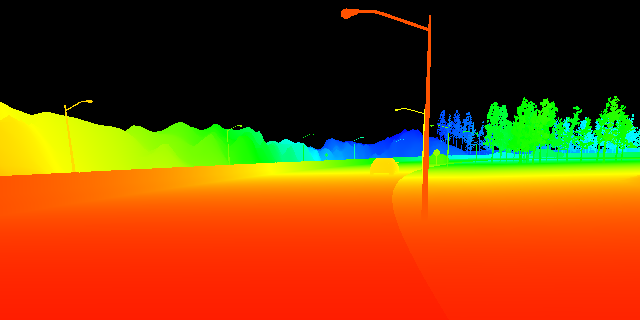}%
		\label{fig:diagram8b}}
	\hspace{0mm}
	\subfloat[]{\includegraphics[width=1.7in]{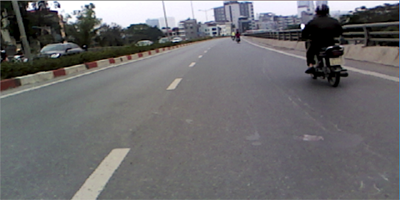}%
		\label{fig:diagram8c}}
	\vspace{0.1in}
	\subfloat[]{\includegraphics[width=1.7in]{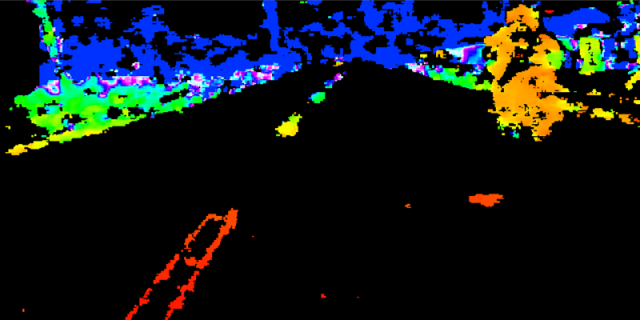}%
		\label{fig:diagram8d}}
	\caption{Differences between images captured (a)-(b) from synthetic data and (c)-(d) by a real RGB-D camera Intel RealSense R200.}
	\label{fig:diagram8}
\end{figure}

\begin{figure}
	\centering
	\subfloat[]{\includegraphics[width=1.7in]{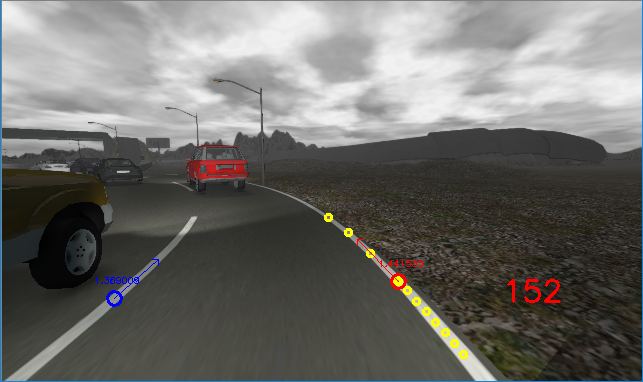}%
		\label{fig:diagram5c}}\vspace{0.1in}
	\subfloat[]{\includegraphics[width=1.7in]{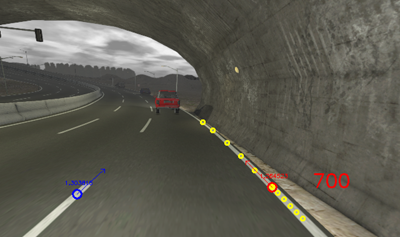}%
		\label{fig:diagram5d}}
	\hspace{0mm}
	\subfloat[]{\includegraphics[width=1.7in]{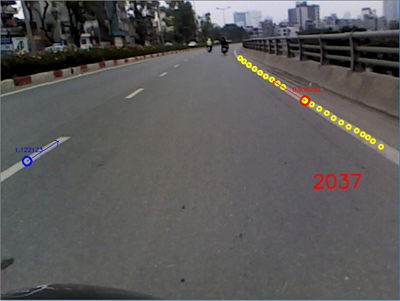}%
		\label{fig:diagram5a}}\vspace{0.1in}
	\subfloat[]{\includegraphics[width=1.7in]{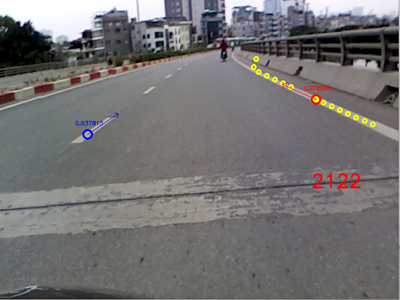}%
		\label{fig:diagram5b}}
	\caption{Detection results with (a)-(b) Synthetic data \\ and (c)-(d) Real data.}
	\label{fig:diagram5}
\end{figure}

Experiments have been conducted with both synthetic and real datasets to evaluate the validity our method under different scenarios and weather conditions. The synthetic datasets are RGB-D images of highway scenario provided by \cite{Ros2016}. The real datasets were recorded by two RGB-D cameras, Microsoft Kinect V2 and Intel Realsense R200. Figure \ref{fig:diagram8} show the differences between images generated by synthetic data and a real RGB-D camera. As shown in Fig.\ref{fig:diagram5}, the data was chosen so that it reflected different road conditions including:
\begin{itemize}
\item	Summer daylight, cloudy and foggy weather.
\item	Lighting changes from overpasses.
\item	Solid-line lane markers.
\item	Segmented-line lane markers.
\item	Shadows from vehicles.
\end{itemize}

In all given conditions, we used templates of $32\times32$ pixels for the colour matching and $5\times5$ window size for normal estimation in FALS. For respond map computation, the depth threshold $T_D$ was set to 20 m based on the range of sensory devices. We chose to use $\alpha =0.4$, $\beta =0.1$, and $\tau_G = 0.5$ to improve the RGB-D respond map. These parameters reflect the contribution of geometric information to the respond map. They are essential to remove the obstacles that cannot be handled by colour template matching. The size of the convolution kernel was $32\times32$ pixels and the minimum jump step $r$ was 5. The condition to activate the template enhancement process is $P_{PCA} > 0.75$. It allows our system to work under moving viewpoint conditions. The camera parameters may affect template's shapes. However, our template size is small to show effects of view perspective.

\begin{figure}
	\centering
	\subfloat[]{\includegraphics[width=1.7in]{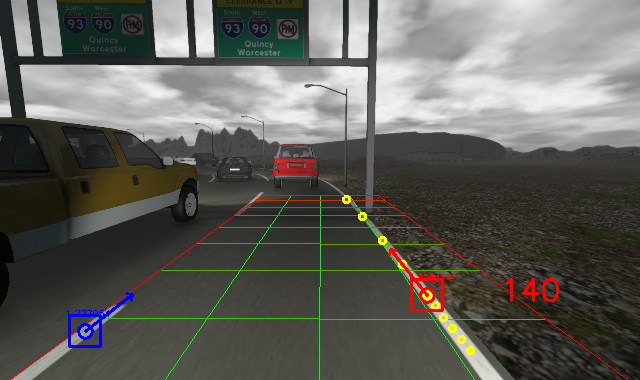}%
		\label{fig:diagram6a}}
	\vspace{0.1in}
	\subfloat[]{\includegraphics[width=1.7in]{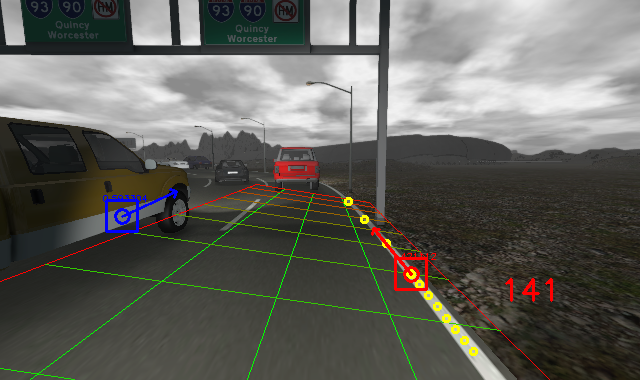}%
		\label{fig:diagram6b}}
	\hspace{0mm}
	\subfloat[]{\includegraphics[width=1.7in]{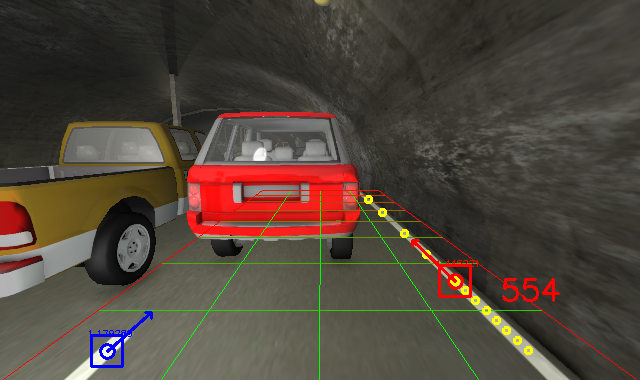}%
		\label{fig:diagram6c}}
	\vspace{0.1in}
	\subfloat[]{\includegraphics[width=1.7in]{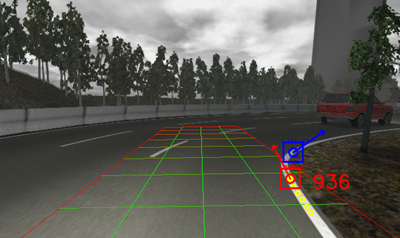}%
		\label{fig:diagram6d}}
		\hspace{0mm}
	\subfloat[]{\includegraphics[width=1.7in]{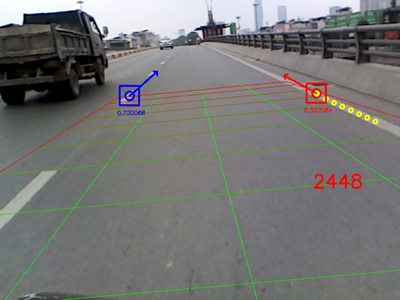}%
		\label{fig:diagram6e}}
	\vspace{0.1in}
	\subfloat[]{\includegraphics[width=1.7in]{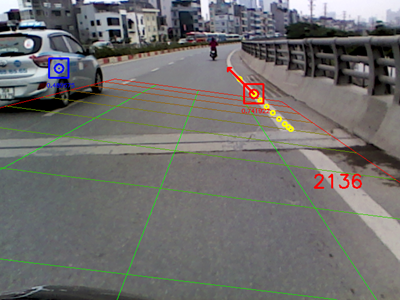}%
		\label{fig:diagram6f}}
		\hspace{0mm}
	\subfloat[]{\includegraphics[width=1.7in]{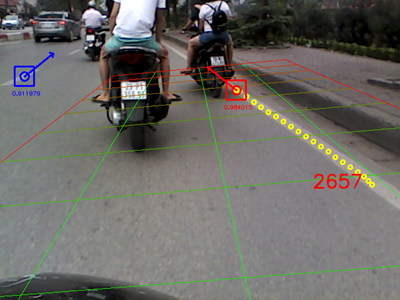}%
		\label{fig:diagram6g}}
	\vspace{0.1in}
	\subfloat[]{\includegraphics[width=1.7in]{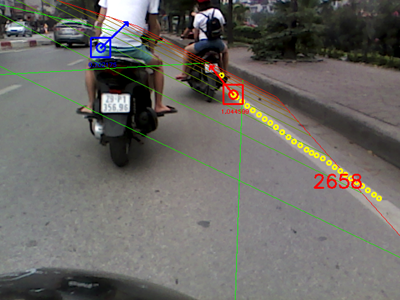}%
		\label{fig:diagram6h}}
	\caption{Detection results with synthetic and real data: (a)-(g) True positive detection; and (b)-(h) False positive detection.}
	\label{fig:diagram6}
\end{figure}

\begin{minipage}{\linewidth}
	\begin{center}
		\bigskip
		\captionof{table}{PERFORMANCE OF 3D LANE MODEL\\ FITTING IN SEVERAL SYNTHETIC AND \\ REALISTIC DATASETS} \label{tab:table1} 
		\begin{tabular}{ | l || c | c |c| p{1cm} }
			\hline
			& \multicolumn{3}{|c|}{Lane detection result}  \\ \hline
			Dataset & Frames & True positive & False positive \\  \hline 
			01-SUMMER & 942 & 87\% & 0.6\% \\ 
			\hline
			01-FOG & 1098 & 84\% & 2\% \\ 
			\hline
			06-FOG & 857 & 80\% & 4\% \\
			\hline
			Our Real Data & 620 & 53\% & 7\% \\
			\hline
		\end{tabular}
		\bigskip
	\end{center}
\end{minipage}

\begin{figure}
	\centering
	\subfloat[]{\includegraphics[width=1.7in]{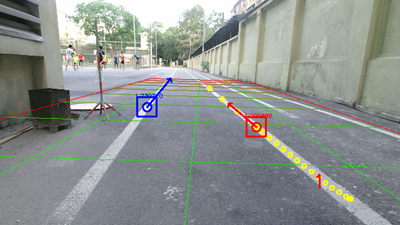}%
		\label{fig:diagram7a}}
	\vspace{0.1in}
	\subfloat[]{\includegraphics[width=1.7in]{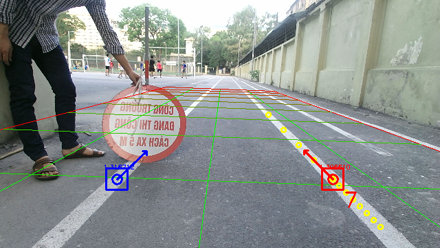}%
		\label{fig:diagram7b}}
	\caption{Detection results with real data recorded by Microsoft Kinect V2 RGB-D camera.}
	\label{fig:diagram7}
\end{figure}

Figures \ref{fig:diagram5} - \ref{fig:diagram7} show the detection results. It can be seen that our method works well for both synthetic datasets and real data captured by the Kinect RGB-D camera. Table 1 shows the performance of our method on synthetic and realistic datasets. Changes in lighting conditions have a little effect on the results. However, wrong detections are sometimes happened, as illustrated in Fig.\ref{fig:diagram5b}, when objects have similar shapes as lane markers. This problem can be improved by using negative filters.

\begin{figure}
	\centering
	\includegraphics[width=9cm]{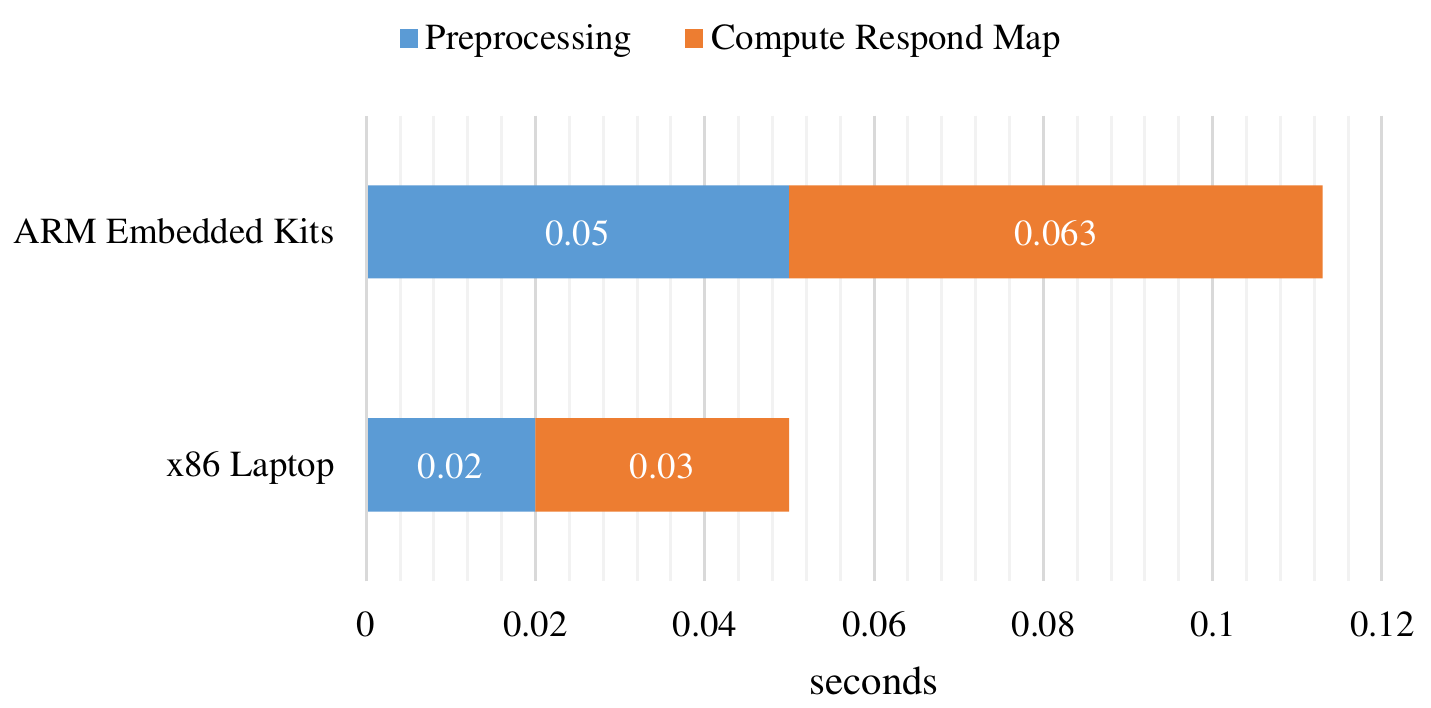}
	\caption{Processing timelines of our algorithm running on laptop and embedded computers.}
	\label{fig:processingtime}
\end{figure}

In experiments with real data recorded by the Realsense R200 RGB-D camera, as shown in Fig.\ref{fig:diagram8b} and Fig.\ref{fig:diagram8d}, low quality and sparse depth data reduce the quality of respond maps causing a number of frames to be skipped. We tried to tune parameters such as $P_{PCA}$ and $\tau_G$ to improve the results, but the false positive rate was also increased. It can be concluded that depth information plays an important role in our detection algorithm. If high-quality devices like Kinect V2 is not available, approaches to interpolate sparse depth images should be considered.

In our implementation, the detection algorithm was written in C++ with OpenCV library and tested in two different hardware platforms: a laptop running Core i7 2.6 GHz CPU and an embedded computer named Jetson TX2 running Quad ARM® A57/2 MB L2. Without using any computation optimisation, the program took around 0.05 seconds on the laptop and 0.113 seconds on the embedded computer to process a single frame. The algorithm is thus feasible for real-time detection. In a further evaluation, the computation time includes nearly 40\% for preprocessing steps and 60\% for computing the respond map (Fig.\ref{fig:processingtime}). Other processing steps require so low computation cost that they do not influence the real-time performance. The cause is numeric operations on large-size matrices. This suggests future works to focus on matrix operation as well as taking advantage of parallel computing techniques such as CUDA with graphical processing units (GPU) for better processing performance.

\section{Conclusion} \label{conclusion}
In this work, we have proposed a new approach to detect lane markers by using a single RGB-D camera. We have also shown that by utilising both colour and depth information in a single processing pipeline, the detection result can be greatly improved with the robustness against illumination changes and obstacle occurrence. In addition, the approach can achieve the real-time performance within a low computational hardware platform with low-cost cameras. It is thus suitable for implementing in various types of vehicle from cars to motorcycles. Our future work will focus on finding the rationale in false-positive scenarios to further improve the detection performance. 

\section*{Acknowledgment}
This work is supported by the grant QG.16.29 of Vietnam National University, Hanoi.

\balance

\bibliographystyle{IEEEtrans}

% that's all folks
\end{document}